\documentclass[letterpaper, 10 pt, conference]{ieeeconf}  

\IEEEoverridecommandlockouts                              

\usepackage{graphics} 
\usepackage{amsmath} 
\usepackage{amssymb}  
\usepackage{subcaption}
\usepackage{graphicx}
\usepackage{paralist}
\usepackage{multirow}
\usepackage{paralist}
\usepackage{dsfont}
\usepackage{float}
\usepackage{makecell}
\usepackage{booktabs}
\usepackage{verbatim}
\usepackage{mathtools}
\usepackage[table]{xcolor}

\definecolor{green}{RGB}{3,112,15}

\newcommand{\mypara}[1]{{\smallskip\noindent \bf #1.}\hspace{0.1in}}

\let\eqref\undefined
\newcommand{\figref}[1]{Fig.~\ref{fig:#1}}

\newcommand{\secref}[1]{Section~\ref{sec:#1}}
\newcommand{\tabref}[1]{Table~\ref{tab:#1}}
\newcommand{\eqref}[1]{Eq.~\ref{eq:#1}}

\newcommand{\figlabel}[1]{\label{fig:#1}}

\newcommand{\seclabel}[1]{\label{sec:#1}}
\newcommand{\tablabel}[1]{\label{tab:#1}}
\newcommand{\eqlabel}[1]{\label{eq:#1}}

\newcommand{\eg}{\textit{e.g.,}}

\newcommand{\vs}{\textit{vs.}}

\newcommand{\ApproachName}{{\textbf{VRL-PAP}}}
\newcommand{\ApproachNameFull}{{\textbf{Visual Representation Learning for Preference-Aware Path Planning}}}

\title{\LARGE \bf
Visual Representation Learning for Preference-Aware Path Planning
}

\author{Kavan Singh Sikand$^{1}$, Sadegh Rabiee$^{1}$, Adam Uccello$^{1,2}$, Xuesu Xiao$^{1}$,  Garrett Warnell$^{1,2}$, Joydeep Biswas$^{1}$
\thanks{$^{1}$ The authors are with the Department of Computer Science, The University of Texas at Austin, Austin, TX. Email: {\tt\small \{kvsikand@cs, srabiee@cs, adam.uccello@austin, xiao@cs, warnellg@cs, joydeepb@cs \}.utexas.edu }}
\thanks{$^{2}$ Adam Uccello and Garrett Warnell are also with the United States Army Research Laboratory (ARL)}
}

\begin{document}

\maketitle
\thispagestyle{empty}
\pagestyle{empty}

\begin{abstract}



Autonomous mobile robots deployed in outdoor environments must reason about different types of terrain for both safety (\eg{} prefer dirt over mud) and deployer preferences (\eg{} prefer dirt path over flower beds). Most existing solutions to this preference-aware path planning problem use semantic segmentation to classify terrain types from camera images, and then ascribe costs to each type. Unfortunately, there are three key limitations of such approaches -- they 1) require pre-enumeration of the discrete terrain types, 2)    are unable to handle hybrid terrain types (\eg{} grassy dirt), and 3) require expensive labelled data to train visual semantic segmentation. We introduce \ApproachNameFull{} (\ApproachName), an alternative approach that overcomes all three limitations: \ApproachName{} leverages unlabelled human demonstrations of navigation to autonomously generate triplets for learning visual representations of terrain that are viewpoint invariant and encode terrain types in a continuous representation space. The learned representations are then used along with the same unlabelled human navigation demonstrations to learn a mapping from the representation space to terrain costs. 
At run time, \ApproachName{} maps from images to representations and then representations to costs to perform preference-aware path planning. 
We present empirical results from challenging outdoor settings that demonstrate \ApproachName{} 1)  is successfully able to pick paths that reflect demonstrated preferences, 2) is comparable in execution to geometric navigation with a highly detailed manually annotated map (without requiring such annotations), 3) is able to generalize to novel terrain types with minimal additional unlabeled demonstrations.

\end{abstract}

\section{Introduction and Related Work}

Autonomous navigation through unstructured human environments is a well-studied problem in robotics, and has seen a number of different approaches. A common class of autonomous navigation is \emph{geometric navigation}, which plans obstacle-free paths purely via geometric collision-checking. Geometric navigation has been shown to be successful over long-term deployments in indoor settings~\cite{biswas2016, hawes2017strands, khandelwal2017bwibots}.

However, geometric navigation is unable to reason about paths over different terrain that may appear to be equally valid geometrically (\eg{} sidewalk \vs{} dirt \vs{} gravel), but have different costs due to reliability of navigation or social norms; or terrain that may seem geometrically impassable but may actually be navigable (\eg{} tall grass). This shortcoming of geometric navigation has motivated a field of research in the space of \emph{visual navigation}, which uses image data from the mobile robot to reason about the environment while navigating.

End-to-end learning solutions to the visual navigation problem, which involve using deep neural networks to learn a policy which predicts control commands given raw sensory inputs, have recently become a field of great interest. The supervised approach to this learning problem is to use a reference policy (usually provided by a human) as the training signal indicating the desired behaviour for a given sensory input~\cite{giusti2015machine}. To avoid the need to provide training labels for every input, Reinforcement Learning (RL) has gained popularity as a method for learning end-to-end policies in a variety of simulation domains~\cite{mnih2013playing} and more recently on real robots~\cite{chiang2019}. 
BADGR~\cite{badgr2021} leverages the available sensing redundancy on a mobile robot to learn affordances of different types of terrain in a self-supervised manner.
By exploring off-policy paths, they are able to learn a planner that ignores geometric obstacles that the robot can safely traverse (\eg{} tall grass). 
In this work, we handle a different case: when geometric information tells us there are \emph{no} obstacles in a given region, but visual information tells us it would be preferable to avoid that region anyway.

While end-to-end approaches are attractive due to their ability to be learned from high-level navigation demonstrations, they have been shown to have significant difficulty generalizing to new environments \cite{xiao2020motion}. To resolve this generalizability issue, a number of approaches start by processing the input to produce some intermediate representation of the environment, such as cost maps, segmentation maps \cite{wigness2018, mousavian2019visual}, or traversability estimates \cite{HiroseSVGS18}, and then perform planning using that data as an input. For example, GoNet~\cite{HiroseSVGS18} uses Generative Adversarial Networks (GANs) to predict the traversability of an environment given nominal examples of navigation for a mobile robot. Because there are a variety of ways of pre-processing visual information which can be useful for different specific downstream navigation tasks, there has also been work focused on choosing between various intermediate representations, and fusing these outputs together before selecting an appropriate action \cite{shen_2019}.

Although intermediate data representations such as semantic segmentation and traversability estimation provide helpful generalizability properties, they often require dense manual labelling to train. This time-intensive process is required for handling any new object or terrain types in the environment.
To ameliorate this shortcoming, inverse reinforcement learning (IRL) with visual semantic segmentation~\cite{wigness2018} learns the navigation cost associated with each semantic class autonomously from human demonstrations. 
A similar approach to learning visual navigation is to frame it as a reinforcement learning problem given semantic segmentation of input images~\cite{pmlr-v87-bruce18a}. While these approaches do not use any explicit labelling when training the navigation planner, they still rely on the outputs of a pre-trained semantic segmentation network, and require manual annotation to extend the semantic segmentation to novel terrain types.

Our approach retains the generalizability benefits of using an intermediate representation while removing the dependence on explicit labelling of visual information. In our approach, both the visual representations and the navigation planner can be adapted to a new environment using only unlabeled human-provided demonstrations.
\section{Preference-Aware Path Planning}
\newcommand{\appearance}{\ensuremath{\phi}}
\newcommand{\appearances}{\ensuremath{\Phi}}
\newcommand{\patches}{\ensuremath{\mathds{I}}}
\newcommand{\patch}{\ensuremath{I}}

We consider the path planning problem in the context of a state space $\mathcal{S}$, action space $\mathcal{A}$, and deterministic transition function $\mathcal{T}: \mathcal{S} \times \mathcal{A} \rightarrow \mathcal{S}$.
Our state space is comprised of states $s = \left[ x, y, \theta, \appearance{} \right] $, where $\left[x,y, \theta\right] \in SE(2)$ denote the robot's position, and $\appearance{}\in \appearances$ denotes the visual appearance of the ground at this location. We define \appearances{} as the space of visual appearances relevant for preference-aware planning.
The action space and transition function are defined by the kinodynamic constraints of the robot.
Given a start state $s_0$ and goal $G$, the local path planning problem is the search for a finite receding horizon sequence of actions, $(a_0, \ldots, a_{N-1}) \in \mathcal{A}^N$ such that the resulting trajectory $\Gamma : i \in \{1,\ldots, N\} \rightarrow \mathcal{S}$, which is defined by $\Gamma(i) = \mathcal{T}(\Gamma(i-1), a_{i-1})$, exhibits minimal cost $J(\Gamma)$, $J: \mathcal{S}^N \rightarrow \mathbb{R^+}$. Since this is a receding horizon local planning problem, the final state of the optimal solution $\Gamma^*(N)$ may not reach $G$, but the trajectory must be optimal with respect to the cost such that $\Gamma^* = \arg_\Gamma\min J(\Gamma)$.

In purely geometric approaches to local planning (i.e., those that consider only geometric obstacles and treat all free space as equal), a common choice for $J$ is
\begin{equation}
    J(\Gamma) = J_f(\Gamma(N), G) + J_l(\Gamma) + J_{cl}(\Gamma) \;,
    \eqlabel{cost_geo}
\end{equation}
where $J_f$ is the cost based on progress towards $G$ (e.g., $J_f(s,G) = ||G - s||$), $J_l$ is the cost based on the free path length of the trajectory, and $J_{cl}$ the cost based on obstacle clearance~\cite{richardson2011} along $\Gamma$. $J_l$ and $J_{cl}$ are computed by comparing $\Gamma$ to the position of observed geometric obstacles in the environment.

Unlike purely geometric approaches, the path planning method we propose seeks to  make preference-aware planning decisions also based on the appearance $\appearance\in\appearances$ of the states of the robot's trajectory. 
For a preference-aware planner that reasons only about distinct semantic classes, \appearances{} would be the set of discrete known semantic classes. In contrast, in our approach $\appearances{}\subset \mathds{R}^k$ is a \emph{continuous space} of k-dimensional learned visual representations relevant for preference-aware planning. To incorporate this visual information into the path planning problem, we add an additional term to \eqref{cost_geo}, redefining $J$ as 
\begin{equation}
    J(\Gamma) = J_f(\Gamma(N), G) + J_l(\Gamma) + J_{cl}(\Gamma) + J_p(\Gamma) \;,
    \label{eqn_cost}
\end{equation} where $J_p(\Gamma)$ computes a cost based on the appearance of the terrain over which the trajectory $\Gamma$ traverses. Intuitively, this cost should be large for trajectories that cause the robot to traverse undesirable terrain, and small otherwise.

Instead of specifying $J_p$ manually, we learn it from human demonstrations that implicitly provide information about terrain desirability using a representation learning approach. In the next section, we will discuss this learning problem.

\section{Visual Representation-Based\\ Preference Learning}

While each robot state $s \in S$ has some true visual appearance $\appearance\in\appearances$, the robot does not have a-priori information about it. Instead, the robot observes image patches of the ground $I \in \mathds{I}$, which are then used to infer $\appearance$ as follows. First, we use an image projection operator $P : S \times S \rightarrow \patches{}$, to identify image patches of the ground in one state as seen by another robot state: $P(s_1,s_2)$ returns the image patch $I_1$ corresponding to the state $s_1$ while observing it from the state $s_2$. Note that $P(s_1,s_2)$ needs access to the full image observation of the robot while at pose $s_2$ -- we assume this image to be available, and omit it from the notation for simplicity. This projection operator can be derived from the camera's extrinsic and intrinsic calibration, and the relative positions of the states $s_1$ and $s_2$. We then apply a visual representation function $f_\mathrm{vis}$ to infer the visual appearance information from this image patch. Thus, the appearance $\phi_1$ of state $s_1$ is inferred via the visual observations from a different state $s_2$ as $\phi_1 = f_\mathrm{vis}(P(s_1, s_2))$.

Given an initial state $s_0$ from which the robot can observe future states along trajectory $\Gamma$, we formulate the preference-aware cost $J_p(\Gamma)$ as
\begin{align}
    \eqlabel{preference_cost}
    J_p(\Gamma) = \sum_{t=0,\ldots,N} \gamma^t 
        J_c(\phi_t), \quad
    \phi_t = f_\mathrm{vis}(P(\Gamma(t),s_0)) ,
\end{align}
 where $f_\mathrm{vis} : \patches{} \rightarrow \appearances{}$ is a visual representation function that maps image patch observations $I \in \patches{}$ to the visual appearance of the ground $\phi \in \appearances{}$, $J_c : \appearances{} \rightarrow \mathbb{R^+}$ is a cost function that uses these embeddings to produce a real-valued cost, and $\gamma^t$ is a discount factor to ensure states far in the future don't have an overbearing impact on the current cost calculation. 
We propose to learn $f_\mathrm{vis}$ via \emph{representation learning}, which has recently shown great success at closing the gap between supervised and unsupervised learning for visual tasks such as image recognition~\cite{Chen2020ASF} and video representation learning~\cite{qian2021spatiotemporal}. We leverage unlabeled human demonstrations to learn the functions $f_\mathrm{vis}$ and $J_p$, as described below.

The training data for this learning problem consists of a set of human-provided demonstrations $D = \left\{\Gamma^D_{i=1:M}\right\}$, where each demonstration $\Gamma^D_i$ consists of a sequence of robot locations and image observations. Each $\Gamma^D_i$ is collected while manually driving a robot from an arbitrary start to a goal location, following paths that encode the human demonstrator's preferences.

\subsection{Visual Representation Learning}
The goal of the visual representation function $f_\mathrm{vis}$ is to map an image patch $I \in \patches{}$ to a low-dimensional representation vector $\appearance{} \in \appearances{}$ that captures only the salient visual information from the patch relevant to preference-aware planning. Ideally, we would like this representation to exhibit two properties: \begin{inparaenum}[1)]
\item separability of patches of terrain with different preference values, and
\item invariance to viewpoint changes (that is, the visual appearance $\appearance{}$ of a given patch of terrain is the same, no matter where it is observed from)\end{inparaenum}. We learn this function using a triplet loss function, a form of contrastive loss \cite{hermans2017defense}, which requires identification of training triplets that simultaneously encourage the desired separability and invariance properties. We next discuss the method for triplet identification as well as the loss function used to learn $f_\mathrm{vis}$.

\mypara{Loss Function}
We define a loss function for $f_\mathrm{vis}$ such that the learned result exhibits the above properties when trained over training triplets collected in a self-supervised procedure. In this loss function, we require a triplet of image patches $\langle I^a, I^s, I^d \rangle$, which are referred to as the anchor, similar, and dissimilar patches respectively.
Our loss function 
\begin{align}
    \eqlabel{loss-vis}
    & L_\mathrm{vis}(f_\mathrm{vis}, \langle I^a, I^s, I^d \rangle) = \\
      &\mkern3mu \max(||f_\mathrm{vis}(I^a) - f_\mathrm{vis}(I^s)|| - ||f_\mathrm{vis}(I^a) - f_\mathrm{vis}(I^d)|| + \delta, 0), \nonumber
\end{align} enforces that in the embedding space $f_\mathrm{vis}(I^a)$ is closer to $f_\mathrm{vis}(I^s)$ than it is to $f_\mathrm{vis}(I^d)$ by at least a fixed margin $\delta$.
Given a training dataset of triplets $R_D = \{\langle I^a_i, I^s_i, I^d_i\rangle_{i=1:N}\}$, the visual representation learning problem finds $f_\mathrm{vis}^*$ which satisfies
\begin{equation}
    f^*_\mathrm{vis} = \arg_{f_\mathrm{vis}} \min \quad \sum_{\mathclap{\langle I^a_i, I^s_i, I^d_i\rangle \in R_D}} L_\mathrm{vis}(f_\mathrm{vis}, \langle I^a, I^s, I^d \rangle).
\end{equation}
Next, we explain the process of obtaining our training dataset $R_D$ from the demonstrated trajectories $D$ such that this loss function will encourage learning representations which satisfy the desired properties given above.

\mypara{Similar Patch Extraction}
In order to enforce viewpoint invariance, we choose triplets such that $I^a$ and $I^s$ are different views of the \textit{same} location in the real world. Viewpoint invariance requires that for all arbitrary states $s,s',s''$, the visual representation of the image patch of the observation of state $s$ should be the same as seen from $s'$ and $s''$:
\begin{align}
    \forall s, s', s''\in S, \quad f_\mathrm{vis}(P(s, s')) = f_\mathrm{vis}(P(s, s'')).
\end{align}
To extract pairs $I^a, I^s$ which correctly enforce this property from the human demonstrations, we use the following procedure. For three arbitrary time-steps $t_1<t_2<t_3$ in a human demonstration trajectory $\Gamma^D_i$, the anchor image patch $I^a$ and similar image patch $I^s$ are selected as the image projections of  $\Gamma^D_i(t_3)$ from time-steps $t_1, t_2$ respectively:
\begin{align}
    I^a = P(\Gamma^D_i(t_3), \Gamma^D_i(t_1)), \quad
    I^s = P(\Gamma^D_i(t_3), \Gamma^D_i(t_2)).
\end{align}
\figref{patch_extraction_sim} illustrates this procedure for similar patch extraction.

\mypara{Dissimilar Patch Extraction}
When selecting patches to identify as dissimilar ($I^d$), we seek to ensure that regions the human demonstrator chose to avoid are distant in the embedding space from regions the demonstration traversed. Rather than ask for human annotation, our approach infers this preference based on the sequence of future demonstration states in each $\Gamma^D_i$. For each demonstration trajectory, the robot generates a hypothesized trajectory $\hat{\Gamma}^D_i$ such that they both start and end at the same points ($\hat{\Gamma}^D_i(0) = \Gamma^D_i(0), \hat{\Gamma}^D_i(N) = \Gamma^D_i(N)$) and the length of the hypothesized path\footnote{We assume the demonstrations are non-trivial, such that $\hat{\Gamma}^D_i$ always exists and has no geometric obstacles, else the corresponding demonstration trajectory $\Gamma^D_i$ is discarded.} is shorter than that of the demonstration: $||\hat{\Gamma}^D_i|| < \Gamma^D_i$. The dissimilar patch $I^d$ is then selected as the image projection of a randomly chosen state on the hypothesized patch that is distant from the demonstration trajectory:
\begin{align}
    I^d &= P(\hat{s}, \Gamma^D_i(t_1)) \\
    \hat{s} &\in \hat{\Gamma}^D_i : \min_t||\hat{s} - \Gamma^D_i(t)|| > T, \nonumber
\end{align}
where $T$ is a tunable threshold. By selecting patches along this hypothesized path which is also far from the demonstrated trajectory, we can infer that this region was explicitly avoided by the human demonstrator. \figref{patch_extraction_diff} provides a visualization of this patch selection procedure.

\begin{figure}
    \vspace{0.5em}
    \centering
    \begin{subfigure}{0.45\textwidth}
     \includegraphics[width=\textwidth]{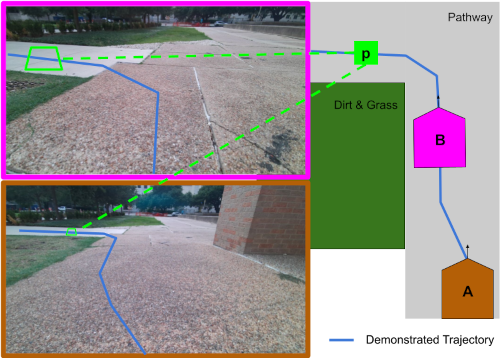}
     \caption{Similar Patch Extraction. The visual representations of patches at location $p$, as observed from the robot at states $A$ and $B$, are enforced to be similar.}
     \figlabel{patch_extraction_sim}
    \end{subfigure}
    \begin{subfigure}{0.45\textwidth}
    \includegraphics[width=\textwidth]{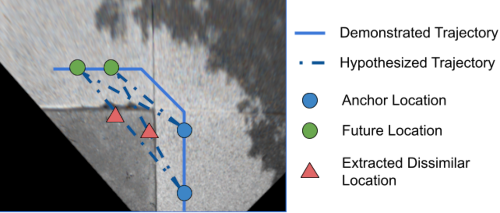}
     \caption{Dissimilar Patch Extraction}
    \figlabel{patch_extraction_diff}
    \end{subfigure}
    \caption{Patch Extraction Procedure}
\end{figure}

\mypara{Triplet Selection}
\seclabel{triplet_selection}
The complete training dataset $R_D = \{\langle I^a_i, I^s_i, I^d_i\rangle_{i=1:N}\}$ is generated in a self-supervised manner by repeating the above procedure over exhaustively chosen time-steps $t_1, t_2, t_3$ for all demonstration trajectories $\Gamma^D_i$ -- since there exists a large number of such time-steps in each demonstration trajectory, we are able to construct a sizable training set with a small number of human demonstrations.

\subsection{Visual Preference Cost Function}
The cost function $J_c : \appearances{} \rightarrow \mathds{R}^+$ is responsible for taking the visual appearance of a single patch of terrain $\appearance{}$, as obtained from $f_\mathrm{vis}$, and outputting a real-valued traversal cost, reflecting the cost incurred by travelling over this terrain. These individual patch costs are then combined together in \eqref{preference_cost} to contribute to the overall cost of a trajectory $\Gamma$.

\mypara{Loss Function}
To train our cost function, we use the same training set that was extracted in \secref{triplet_selection}. Our loss function has a margin $\delta_c$, and can be defined as:
\begin{align}
\begin{split}
L_c(J_c, \phi^p, \phi^n) &= \max(J_c(\phi^p) - J_c(\phi^n) + \delta_c, 0). \\
\end{split}
\end{align}
This loss function enforces that $J_c(\phi^n)$ is at least $\delta_c$ greater than $J_c(\phi^p)$. We therefore want to choose $\phi^n$ such that it is a patch of terrain that should have a high cost relative to  $\phi^p$. To do this, we find the $J_c^*$ which optimizes:
\begin{align}
 J^*_c = \arg_{J_c}\min \quad \sum_{\mathclap{\langle I^a_i, I^s_i, I^d_i\rangle \in R_D}}  &L_c(J_c, f_\mathrm{vis}(I^a), f_\mathrm{vis}(I^d)) \nonumber \\ + &L_c(J_c, f_\mathrm{vis}(I^s), f_\mathrm{vis}(I^d)).
\end{align}
Here, we use $\{ I^a, I^s \}$, which represent image patches over which the robot traversed during demonstration, as the patches which generate $\phi^p$, and we use $\{I^d\}$, which represent image patches that were explicitly avoided during demonstration, as the patches which generate $\phi^n$, thereby extracting the appropriate cost preferences from the demonstration.
By comparing the produced costs in a pairwise fashion, we can enforce a strict ordering among the terrain types that are present. From the demonstrations, we are unable to directly label the correct cost for a region, but rather are given relative preference information, and therefore we do not use a direct regression-based cost function.

\section{Implementation Details}

In this section we discuss details of our implementation that allowed the method described above to be deployed in real-time during our experimental evaluation.

\mypara{Ground-Plane Homography} When working with the robot's visual data, we first undistort the images and apply a homography to transform the images to overhead views, determined by the intrinsics and extrinsics of the robot's camera. \figref{patch_extraction_diff} demonstrates the result of this homography. Since the transformed image has a constant scale with respect to distances in the world, rectangular image patches of constant size correspond to constant size rectangular regions on the ground plane. In our implementation, we chose a patch size of $40 \times 40$ pixels, representing approximately $0.3m^2$ in the real world, comparable to the size of our robot.

\mypara{Local Cost-Map} We persist the costs for each observed patch (the output of $J_c$) in a local costmap centered on the robot's current position, using the robot's odometry estimate to transform the existing costmap between time-steps. This affords the robot a short-term memory of visual information it has observed, but which is no longer in its view, which helps our implementation handle sharp turns and narrow field-of-view cameras. We recompute $J_c$ for any patches which can be observed by the robot at the current time-step, and we recompute $J_p$ from this costmap for each trajectory at every time step.

\mypara{Network Structure} In our implementation, the visual representation function $f_\mathrm{vis}$ takes the form of a neural network with  2 convolutional layers followed by a 3 densely connected layers, and our representation \appearance{} is a 6-dimensional vector. The cost function $J_c$ is a small 3-layer Multi-Layer Perceptron (MLP) with a ReLU activation function to prevent negative outputs.

\mypara{Batched Cost Computation} Because our formulation computes costs for each image patch independently, we are able to parallelize the computation of patch costs for each image. The small patch size combined with the compact network structure described above allows our algorithm to process hundreds of patches per time-step on our robot's modest GPU (Nvidia GeForce GTX 1050TI), which is enough to process an entire image observation. Our processing of visual information occurs at $20 \texttt{Hz}$ during the planning process; significantly faster than FCHarDNet~\cite{Chao2019HarDNetAL}, a segmentation network designed for efficiency in compute-constrained environments, which was only able achieve ${\sim}6 \texttt{Hz}$ when running on the same GPU as part of an autonomy stack.

\begin{figure*}[ht]
\vspace{0.5em}
\centering
\begin{subfigure}[t]{0.6\textwidth}
\includegraphics[width=\textwidth]{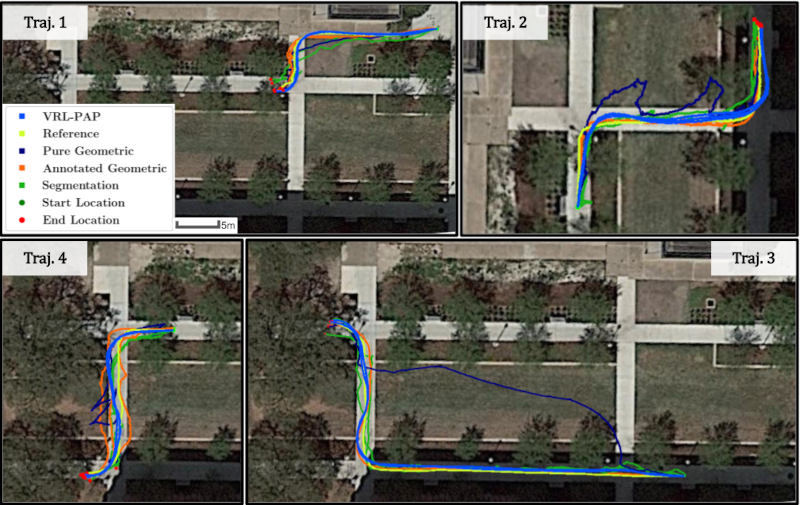}
\caption{Primary Evaluation Environment Trajectories}
\figlabel{evaluation_trajectories_primary}
\end{subfigure}
\begin{subfigure}[t]{0.37\textwidth}
\centering
\includegraphics[width=\textwidth]{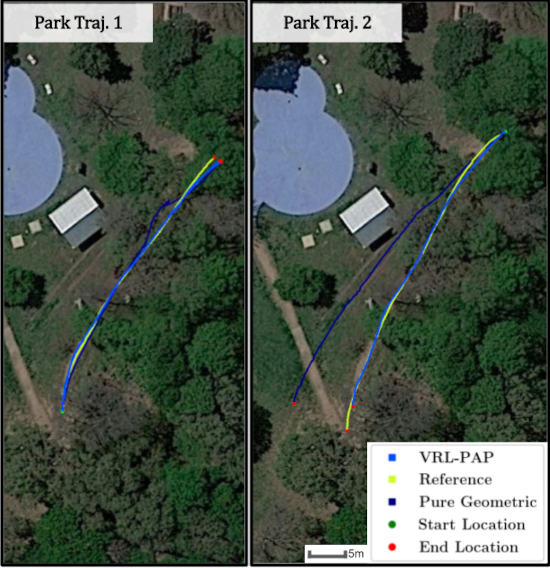}
\caption{Secondary Evaluation Environment Trajectories}
\figlabel{evaluation_trajectories_park}
\end{subfigure}
\caption{}
\end{figure*}

\begin{table*}
\resizebox{\textwidth}{!}{
\begin{tabular}{@{}lcccccccccccc@{}}
\toprule
\multicolumn{1}{c}{} & \multicolumn{3}{c}{\textbf{Trajectory 1}} & \multicolumn{3}{c}{\textbf{Trajectory 2}} & \multicolumn{3}{c}{\textbf{Trajectory 3}} & \multicolumn{3}{c}{\textbf{Trajectory 4}} \\ \cmidrule(lr){2-4} \cmidrule(lr){5-7} \cmidrule(lr){8-10} \cmidrule(lr){11-13}
\multicolumn{1}{c}{Planner} & \begin{tabular}[c]{@{}c@{}}Hausdorff\\ Distance ($m$)\end{tabular} & \begin{tabular}[c]{@{}c@{}}Off-Path\\ Duration ($s$)\end{tabular} &\begin{tabular}[c]{@{}c@{}}Intervention\\ Count\end{tabular} & \begin{tabular}[c]{@{}c@{}}Hausdorff\\ Distance ($m$)\end{tabular} & \begin{tabular}[c]{@{}c@{}}Off-Path\\ Duration ($s$)\end{tabular} & \begin{tabular}[c]{@{}c@{}}Intervention\\ Count\end{tabular} & \begin{tabular}[c]{@{}c@{}}Hausdorff\\ Distance ($m$)\end{tabular} & \begin{tabular}[c]{@{}c@{}}Off-Path\\ Duration ($s$)\end{tabular} & \begin{tabular}[c]{@{}c@{}}Intervention\\ Count\end{tabular} & \begin{tabular}[c]{@{}c@{}}Hausdorff\\ Distance ($m$)\end{tabular} & \begin{tabular}[c]{@{}c@{}}Off-Path\\ Duration ($s$)\end{tabular} & \begin{tabular}[c]{@{}c@{}}Intervention\\ Count\end{tabular} \\ \midrule
\textbf{Preference Learning} & \textbf{0.95} & \textbf{0} & \textbf{0} & \textbf{0.93} & \textbf{0} & \textbf{0} & 1.37 & \textbf{0} & \textbf{0} & \textbf{1.49} & \textbf{0} & \textbf{0} \\
\textbf{Segmentation} & 2.40 & 13 & 3.5 & 1.56 & 2.5 & 2 & 2.23 & 8.5 & 4 & 2.66 & 5.5 & 3 \\
\textbf{Annotated Geometric} & 1.05 & 1.25 & 1 & 1.15 & 0.25 & 0.5 & \textbf{1.32} & \textbf{0} & 1 & 2.17 & 4 & \textbf{0} \\
\textbf{Pure Geometric} & $>1.83\footnotemark[2]$ & 24 & 4 & $>4.56\footnotemark[2]$ & 18 & 5 & $>10.02\footnotemark[2]$ & 62 & 7 & $>4.01\footnotemark[2]$ & 15 & 5 \\ \bottomrule
\end{tabular}
}
\caption{Mean Metrics in Primary Evaluation Environment.}
\tablabel{metrics}
\end{table*}

\section{Experimental Results}

We evaluate \ApproachName{} in a variety of real-world environments by measuring its
\begin{inparaenum}[1)]
\item \emph{accuracy} at following desired paths compared to other visual and geometric navigation planners;
\item \emph{adaptability} to novel terrain types from limited unlabeled demonstration; and 
\item \emph{scalability} to long trajectories in the real world.
\end{inparaenum}
All experiments were performed on a Clearpath Jackal Unmanned Ground Vehicle equipped with a VLP-16 LiDAR, a Microsoft Azure Kinect RGB-D camera, and an Nvidia GeForce GTX 1050TI.

\subsection{Accuracy In Following Desired Paths}
\noindent
We compare \ApproachName\ to a reference and four baselines:
\\\noindent
 \textbf{Reference:} A reference trajectory of the correct preference-aware path provided via joystick by a human operator.
\\\noindent
\textbf{Annotated Geometric:} A geometric planner using a detailed hand-annotated navigation graph of the evaluation environment including desirable paths.
\\\noindent
\textbf{Pure Geometric:} A purely geometric planner without a detailed hand-annotated navigation graph of desirable paths.
\\\noindent
\textbf{Segmentation:} A state of the art preference-aware planner using semantic segmentation to build a local cost map for planning: The Army Research Laboratory's Autonomy Stack, which uses FCHarDNet~\cite{Chao2019HarDNetAL} for semantic segmentation, trained on the RUGD Vision dataset~\cite{RUGD2019IROS}.

We ran repeated trials with each of these navigation methods on four evaluation trajectories. \figref{evaluation_trajectories_primary} shows these trajectories, which traverse a real-world environment that includes multiple types of valid sidewalk, shadows cast by trees and buildings, and multiple types of undesired terrain including dirt, grass, and shrubs. This environment was also used to collect training data for the preference learning model, though the demonstration trajectories were distinct from the evaluation trajectories. 

We use an undirected Hausdorff distance, which measures the distance from each point in the trajectory to the closest point in the reference trajectory, to quantitatively evaluate the accuracy of each autonomously executed trajectory:
\begin{align}
    H(\Gamma_a, \Gamma_b) &= \max (h(\Gamma_a, \Gamma_b), h(\Gamma_b, \Gamma_a)), \\
    h &= \sum_{a \in \Gamma_a}  \min_{b \in \Gamma_b} ||a - b|| . \nonumber
\end{align}
Additionally, we evaluate the duration of time for which the robot was on undesirable terrain type, and the number of operator interventions necessary to prevent the robot from taking unsafe actions (\eg{} driving into dense shrubbery).

The results of these experiments are presented in \tabref{metrics}. From these results, we see that \ApproachName{} performs comparably to \textbf{Annotated Geometric} baseline, without access to the hand-made navigation graph for this environment. Further, \ApproachName{} never needed human intervention to stay on the desired path while all of the baseline approaches required human intervention. The pre-trained segmentation-based approach required interventions due to instances of terrain in the environment that were not in its training dataset (short shrubbery and smooth dirt), which motivates the adaptability experiment presented in \secref{adaptability_eval}.

\footnotetext[2]{Error would have been higher; included extensive manual intervention}

\begin{figure*}[ht]
\vspace{0.5em}
\centering
\begin{subfigure}{0.23\textwidth}
\centering
\includegraphics[width=\textwidth]{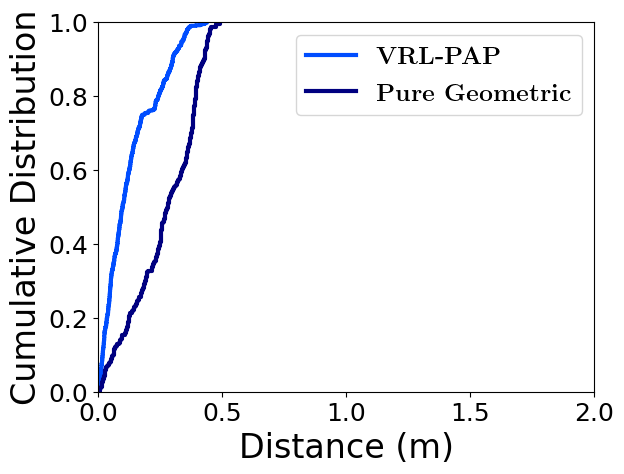}
\caption{Park Traj. 1}
\figlabel{park_disturbance_1}
\end{subfigure}
\begin{subfigure}{0.23\textwidth}
\centering
\includegraphics[width=\textwidth]{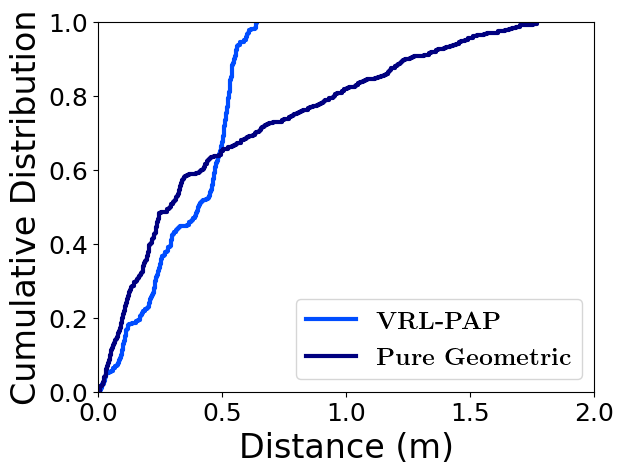}
\caption{Park Traj. 2}
\figlabel{park_disturbance_2}
\end{subfigure}
\begin{subfigure}{0.23\textwidth}
    \centering
    \includegraphics[width=\textwidth]{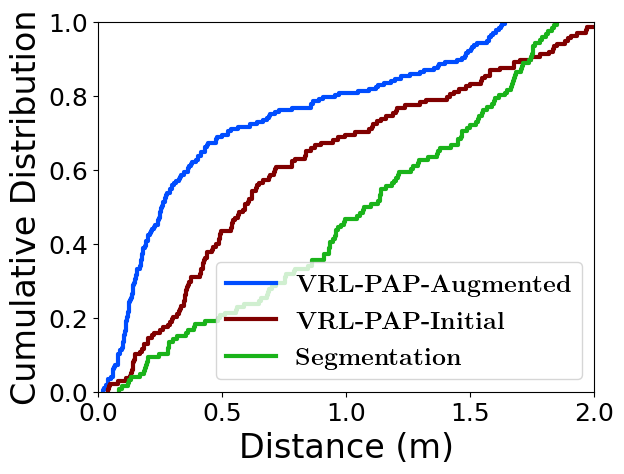}
    \caption{Adaptability Experiment}
    \figlabel{transfer_trajectory_disturbance}
\end{subfigure}
\begin{subfigure}{0.23\textwidth}
\centering
\includegraphics[width=\textwidth]{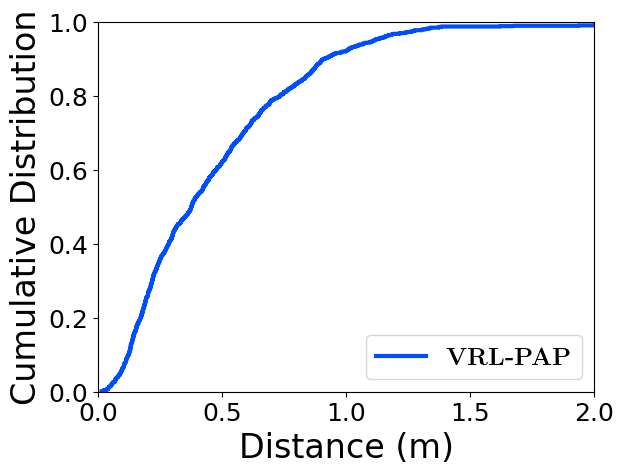}
\caption{Extended Deployment}
\figlabel{long_cdf}
\end{subfigure}
\caption{Cumulative Distribution Functions measuring distance to Reference trajectory}
\end{figure*}

\subsection{Accuracy In Secondary Environment}
\seclabel{complex_env} To investigate the accuracy of \ApproachName{} in a more complex scenario, we performed evaluation in an unstructured park environment, which included paths that were less clearly delineated than those in the primary evaluation environment. \figref{park_path} shows a sample image from this environment, along with a top-down view of the local costmap $J_c$ generated by \ApproachName{} after providing a handful of demonstrations in this environment. \figref{evaluation_trajectories_park} shows the two trajectories over which we evaluated \ApproachName{} and the \textbf{Pure Geometric} baseline, performing two trials of each. \figref{park_disturbance_1} and \figref{park_disturbance_2} show the cumulative distribution of the distance (CDF) from the reference trajectory when executing \ApproachName{} and the \textbf{Pure Geometric} baseline
 -- in both cases \ApproachName{} more closely follows the reference trajectory, staying within $0.75m$ of the reference at all times. 

\begin{figure}[H]
\centering
\includegraphics[width=0.45\textwidth]{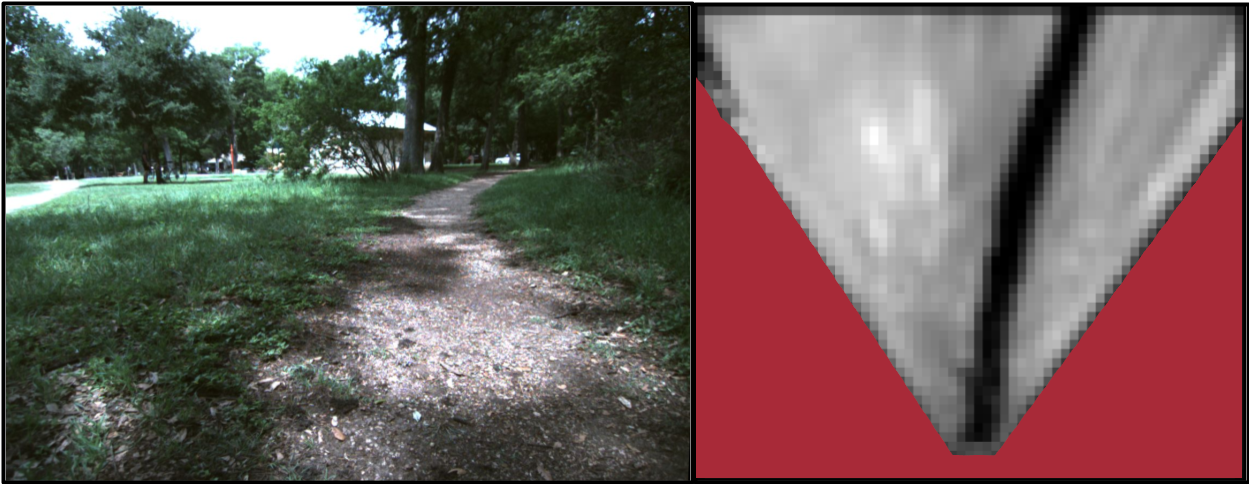}
\caption{Path and top-down costmap $J_c$ in the secondary evaluation environment. Lower cost regions are darker. Red regions are out of the robot's field of view.}
\figlabel{park_path}
\end{figure}

\subsection{Adaptability to Novel Terrain Types}
\seclabel{adaptability_eval}
In principle, both visual segmentation and \ApproachName{} should be customizable to novel terrain types with sufficient training data. However, a key feature of \ApproachName{} is that it can adapt to novel terrain types given only unlabeled human demonstrations. In contrast, for a visual semantic segmentation-based approach, adapting to new types of terrain requires collecting labelled segmentation images, which is significantly more onerous.

To evaluate the adaptability of \ApproachName{}, we first deployed it in an environment with a new class of undesirable terrain.
The initial model \ApproachName\textbf{-initial} failed to avoid the terrain type, as did the segmentation-based approach. However, after providing 3 unlabeled human demonstrations of a different trajectory in the new environment, totalling just 47 seconds of driving, our updated model \ApproachName\textbf{-augmented} was able to successfully avoid the undesirable terrain.
\figref{transfer_trajectory_disturbance} and \figref{transfer_trajectory} show the results of this experiment.

\begin{figure}[h]
    \centering
    \includegraphics[width=0.45\textwidth]{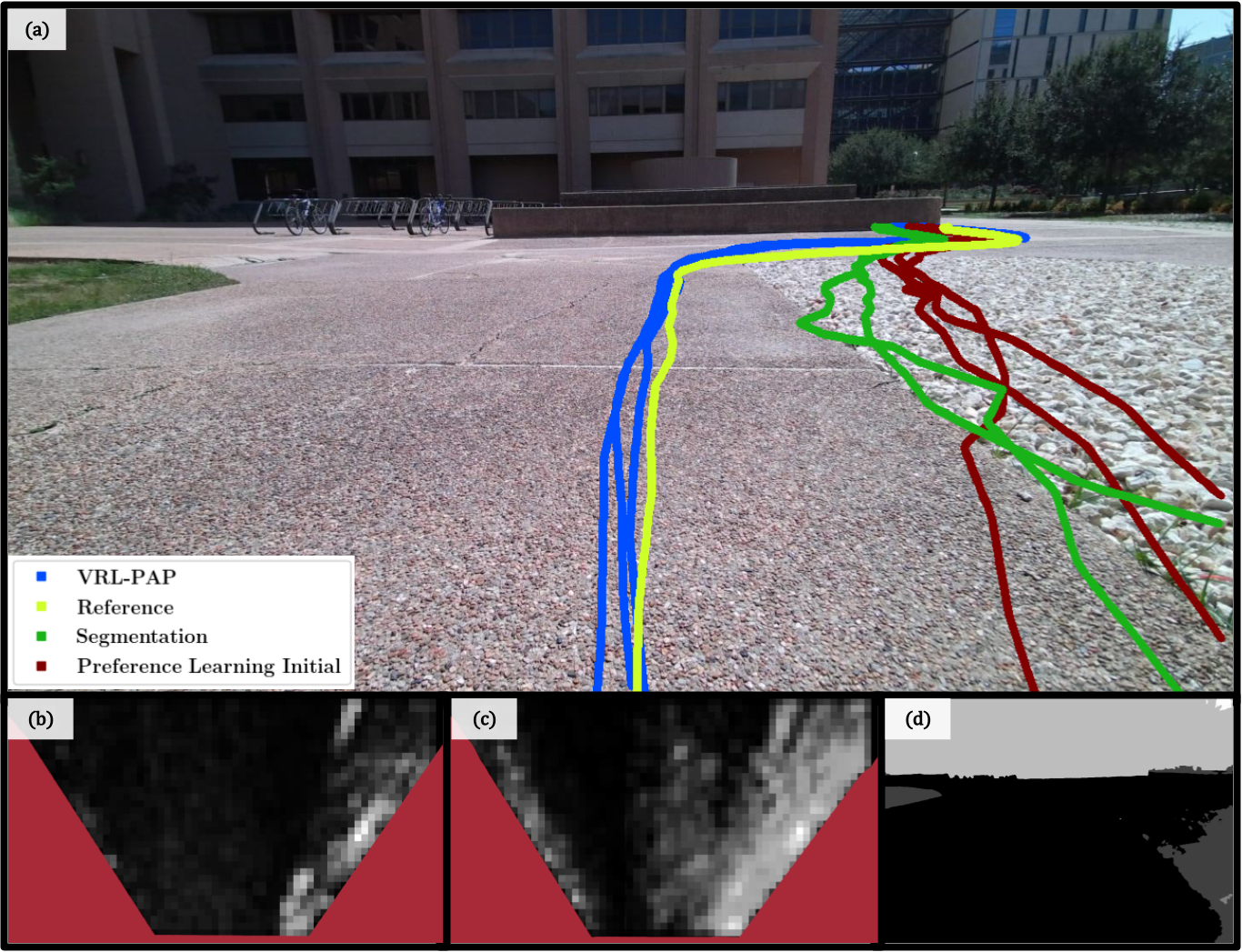}
    \caption{
    Comparison of the behavior of path planners in the presence of a novel undesirable type of terrain (gravel). (a) Visualization of the trajectories traversed by each planner. (b) Top-down view of the predicted terrain cost by \ApproachName{}\textbf{-initial} before being provided any training data including gravel, and (c) by  \ApproachName{}\textbf{-augmented} after re-training given a few short human demonstrations. Darker regions indicate lower navigation cost. Red regions are out of the robot's field of view. (d) Camera-frame view of the predicted cost map by the semantic segmentation approach.}
    \figlabel{transfer_trajectory}
\end{figure}

\subsection{Scalability To Extended Deployments}
Finally, we provide an example of \ApproachName{} navigating a long trajectory, demonstrating its ability to stay on desirable paths over a long period of time without human intervention. \figref{park_long} shows the evaluation trajectory -- it circumnavigates the park environment used in \secref{complex_env}. This trajectory covers $440m$ of autonomous navigation, during which the robot was provided $4$ sequential navigation goals, and required $0$ manual interventions. \figref{long_cdf} shows a CDF of the distance between \ApproachName{} and the human-provided reference trajectory, showing that it stayed less than $1m$ away from the reference for over $90\%$ of the trajectory.

\begin{figure}[h]
\centering
\includegraphics[width=0.45\textwidth]{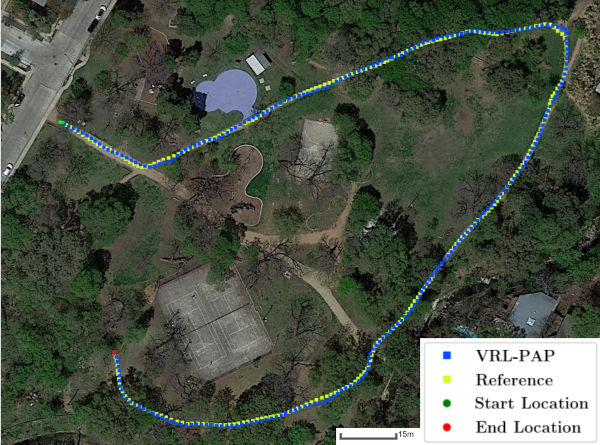}
\caption{Extended Evaluation Trajectory}
\figlabel{park_long}
\end{figure}


\section{Conclusion}

In this work we presented \ApproachName{}, a method for preference-aware path planning based on visual representations, which is learned from unlabelled human demonstrations. We provided a formulation for this approach which enforces desired properties of viewpoint invariance and separability on the learned visual representations. Finally, we demonstrated this approach's capacity to successfully navigate in a variety of environments, as well as transfer to novel terrain types with no manual annotation of training data.

\section{Acknowledgements}
This work has taken place in the Autonomous Mobile
Robotics Laboratory (AMRL) at UT Austin. AMRL research is supported in part by NSF (CAREER-2046955,
IIS-1954778, SHF-2006404), ARO (W911NF-19-2-0333),
DARPA (HR001120C0031), Amazon, JP Morgan, and
Northrop Grumman Mission Systems. The views and conclusions contained in this document are those of the authors
alone.

\addtolength{\textheight}{-12cm}   






\bibliographystyle{IEEEtran}
\bibliography{references} 

\end{document}